\documentclass{ws-ijwmip}
\usepackage{graphicx}
\usepackage{diagbox}
\usepackage[super]{cite}
\usepackage[T1]{fontenc}
\usepackage{multirow}
\usepackage{makecell}
\usepackage{hyperref}
\usepackage{subfig}

\begin{document}

\markboth{S. Yu et al.}{Texture classification network integrating adaptive wavelet transform.}

\catchline{}{}{}{}{}

\title{TEXTURE CLASSIFICATION NETWORK INTEGRATING ADAPTIVE WAVELET TRANSFORM}

\author{SU-XI YU$^\ast$, JING-YUAN HE$^\dagger$ and YI WANG$^\ddagger$}
\address{College of Computer Science\\
	Chongqing University, Chongqing 400030, China\\
	$^\ast$ysx@cqu.edu.cn\\
	$^\dagger$ibm$\_$hjy@cqu.edu.cn\\
	$^\ddagger$yiwang@cqu.edu.cn}

\author{YU-JIAO CAI$^\ast$ and JUN YANG$^\dagger$}
\address{$^\ast$Department of General Surgery, $^\dagger$Department of Ultrasound\\
	Xinqiao Hospital, Chongqing 400037, China\\
	$^\ast$cyj760628@126.com\\
	$^\dagger$woshiyangjun123@126.com}

\author{BO LIN$^\ast$, WEI-BIN YANG$^\dagger$ and JIAN RUAN$^\ddagger$} 
\address{Intelligent Oncology Research Center\\
	Chongqing University Cancer
	Hospital, Chongqing 400000, China\\
	$^\ast$linbo@cqu.edu.cn\\
	$^\dagger$ywb@cqu.edu.cn\\
	$^\ddagger$ruanjian@cqu.edu.cn}

\maketitle

\begin{history}
\received{(Day Month Year)}
\revised{(Day Month Year)}
\accepted{(Day Month Year)}
\published{(Day Month Year)}
\end{history}

\begin{abstract}
Graves' disease is a common condition that is diagnosed clinically by determining the smoothness of the thyroid texture and its morphology in ultrasound images. Currently, the most widely used approach for the automated diagnosis of Graves' disease utilizes Convolutional Neural Networks (CNNs) for both feature extraction and classification. However, these methods demonstrate limited efficacy in capturing texture features. Given the high capacity of wavelets in describing texture features, this research integrates learnable wavelet modules utilizing the Lifting Scheme into CNNs and incorporates a parallel wavelet branch into the ResNet18 model to enhance texture feature extraction. Our model can analyze texture features in spatial and frequency domains simultaneously, leading to optimized classification accuracy. We conducted experiments on collected ultrasound datasets and publicly available natural image texture datasets, our proposed network achieved 97.27\% accuracy and 95.60\% recall on ultrasound datasets, 60.765\% accuracy on natural image texture datasets, surpassing the accuracy of ResNet and confirming the effectiveness of our approach.
\end{abstract}

\keywords{Ultrasound; Graves' disease; Deep Learning; Wavelets Transform; Lifting Scheme.}

\ccode{AMS Subject Classification: 22E46, 53C35, 57S20}

\section{Introduction}\label{Introduction}

Graves' disease is a common thyroid disorder characterized by glandular fibrosis on the surface of the gland. Early diagnosis is crucial for thyroid disease treatment. The advancement of medical technology has introduced low-cost, noninvasive ultrasonography as one of the primary methods for detecting diseases. However, varying use of ultrasonography equipment by technicians can result in inconsistencies in ultrasound image brightness and contrast. Consequently, differing diagnostic interpretations among clinicians may arise. Therefore, the development of an objective and accurate auxiliary tool is highly significant in improving ultrasound clinical diagnosis accuracy.

\begin{figure}[htbp]
	\centerline{\includegraphics[width=10cm]{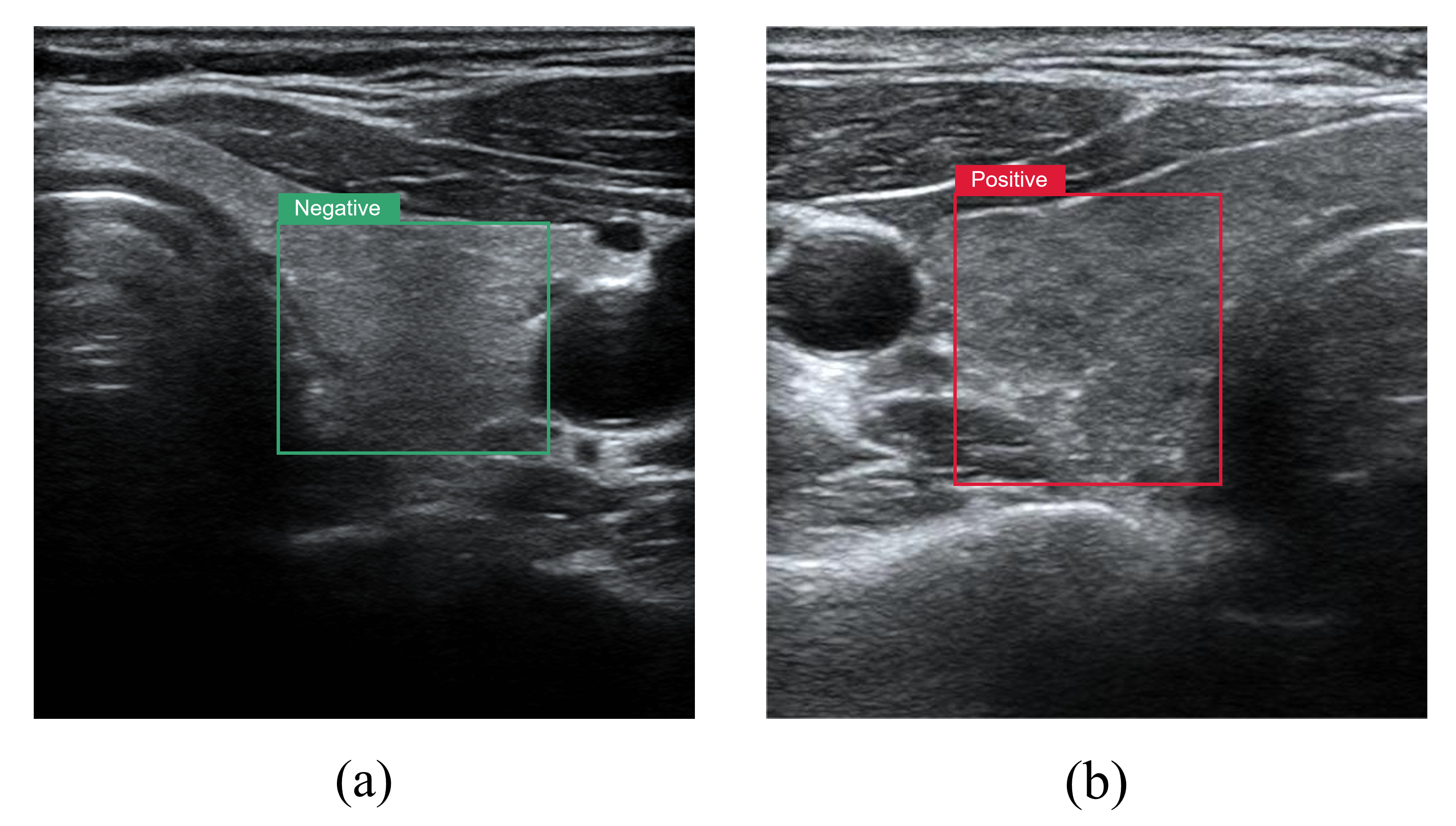}}
	\vspace*{8pt}
	\caption{(a) Ultrasound image of a normal thyroid; (b) Ultrasound image of Graves' disease.}
	\label{fig:1}
\end{figure}

Fig. \ref{fig:1}.(a) illustrates that a normal thyroid gland appears with uniform echogenicity, gray scale and a smooth surface in the ultrasound image, while Fig. \ref{fig:1}.(b) shows that Graves' disease presents with uneven echogenicity and a rough surface, making texture determination key for its diagnosis. Despite the remarkable success of computer-aided analysis techniques facilitating the ultrasound texture diagnosis,\cite{7,10,15} only a few studies have investigated Graves' disease, and most existing works on its diagnosis utilize only traditional or deep learning methods, which inherently possess certain drawbacks.

Traditional methods, such as wavelet transform and fractal dimension, depend heavily on manual extraction and selection of features. Their parameter selection process requires previous experience and expertise; it can be time-consuming and computationally expensive to identify the optimal parameters.\cite{9} Furthermore, recent research indicates that deep learning methods may underperform traditional methods in texture analysis tasks.\cite{2,3,8} To address the drawbacks of these two techniques, this paper proposes a novel approach that combines CNN and traditional texture analysis methods for efficient texture feature extraction from ultrasound images.

To enhance the flexibility and automation of traditional methods and the feature learning process, some scholars have incorporated traditional texture analysis methods into neural networks in a learnable layer-wise manner, such as histogram (see Ref.~\refcite{9}) and fractal dimension (see Ref.~\refcite{14}) , achieving higher accuracy than backbone methods. For wavelet transform, Maria et al.\cite{1} proposed Deep Adaptive Wavelet Network (DAWN) combining Lifting Scheme, which is the first trainable wavelet filter proposed in the context of CNN. However, DAWN only concatenates wavelet modules after a shallow CNN with depth of 2 and does not fully exploit the spatial domain features that can be learned by CNN.

In our network, based on the 2D Adaptive Lifting Scheme in Ref.~\refcite{1}, the Haar discrete wavelet transform is used to achieve the splitting of the signal,\cite{13} which in turn reduces the number of Lifting Scheme modules required to achieve one level of wavelet decomposition of a 2D image from three to one. Meanwhile, unlike DAWN, the network in this paper combines a cluster of wavelet transform modules integrated with multi-resolution analysis in parallel with CNN, enabling the classifier to perform diagnosis based on features in both spatial and frequency domains.

In this paper, a CNN integrated with a learnable wavelet transform is applied to ultrasound diagnosis of Graves' disease. The proposed framework is evaluated on a collected data set and it demonstrates significant improvement in accuracy and recall over the advanced neural network for texture classification.

The contributions of this work are:
\begin{itemlist}
	\item A learnable wavelet transform module is designed based on Lifting Scheme, which internally employs discrete wavelet transform for signal splitting and can achieve one level of wavelet decomposition of an image with a single Lifting Scheme module.
	\item Parallelizing a learnable wavelet branch integrated with multi-resolution analysis into a CNN, which allows the network to learn features in both spatial and frequency domains.
	\item The application of CNN incorporated with wavelet transform to ultrasound diagnosis of Graves' disease for high accuracy texture classification.
\end{itemlist}

\section{Method}\label{Method}
The next two subsections will introduce the design of wavelet modules and networks in DAWN and this paper's network, respectively.

\subsection{Design in DAWN}\label{Design in DAWN}

\subsubsection{2D adaptive lifting scheme}\label{2D adaptive lifting scheme}
\begin{figure}[htbp]
	\centerline{\includegraphics[width=13cm]{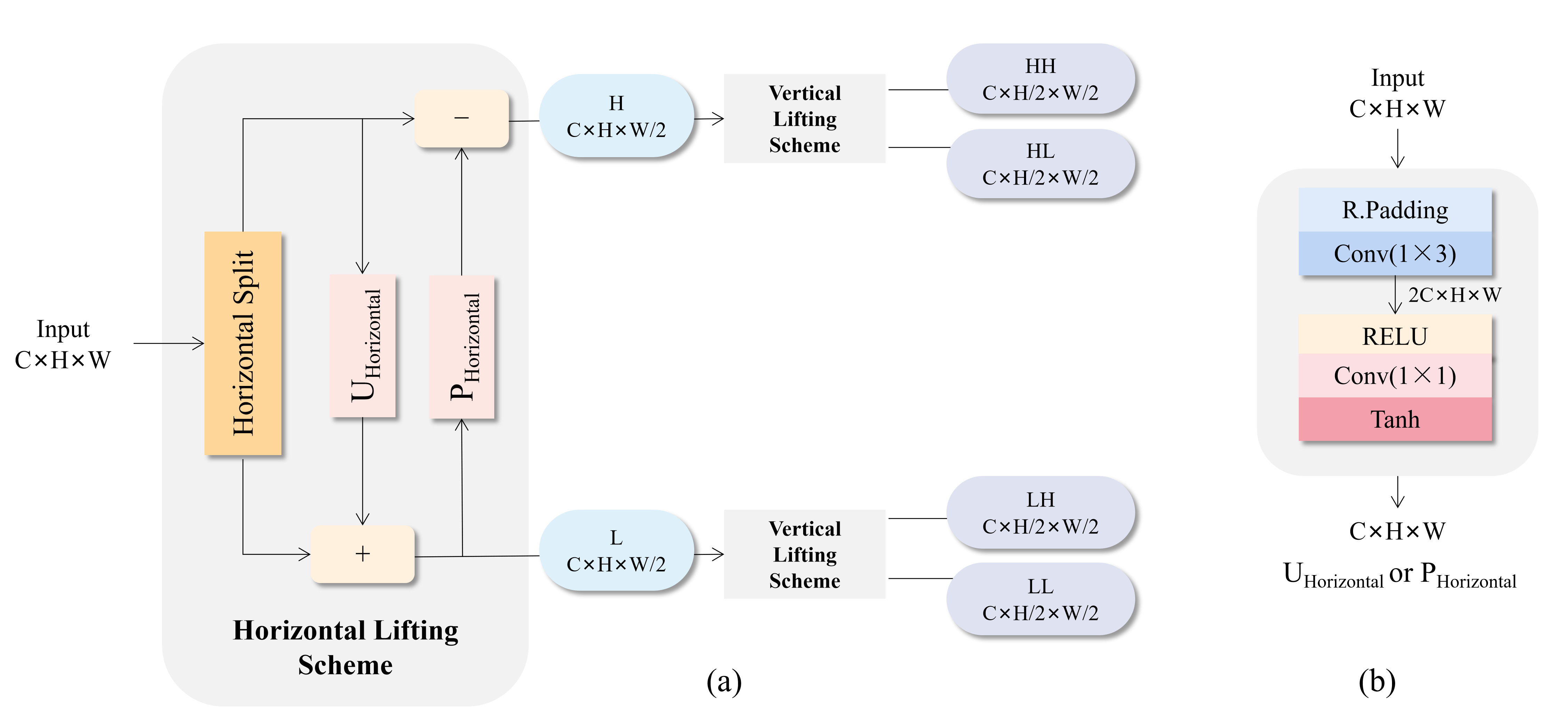}}
	\vspace*{8pt}
	\caption{Structure of 2D Adaptive Lifting Scheme in DAWN.}
	\label{fig:2}
\end{figure}

Lifting Scheme,\cite{12} also known as second generation wavelets, is a simple and powerful method for constructing wavelets, which is divided into three steps, Split, Update, and Predict. Maria et al.\cite{1} proposed a 2D Adaptive Lifting Scheme based on Lifting Scheme, whose structure is shown in Fig. \ref{fig:2}. This module decomposes an image using one horizontal Lifting Scheme and two independent vertical Lifting Schemes, resulting in four components with half the width and height. Each Lifting Scheme module comprises the following three stages.

\textbf{Split} Lazy wavelet transform is used in Ref.~\refcite{1} to decompose the signal $x$ into non-overlapping odd component $x_{o}$ and even component $x_{e}$ in the horizontal or vertical direction. ``(\ref{eq:2.1})'' provide the definition of such decomposition. Despite the ease of implementation, this method can only perform signal decomposition in one direction. Therefore, three Lifting Scheme modules are needed to complete one wavelet decomposition of the image. It is necessary to streamline the architecture design.
\begin{equation}\label{eq:2.1}
x_{o}[n]=x[2n+1], x_{e}[n]=x[2n].
\end{equation}

\textbf{Update and Predict} According to the theory proposed in Ref.~\refcite{16}, Lifting Scheme for one-dimensional signals can be implemented using Back Propagation (BP) Networks, Ref.~\refcite{1} designed predictor and updater as shown in Fig. \ref{fig:2}.(b), where the processing direction of the signal is controlled by the convolution kernel of the first convolution layer. The horizontal direction is processed at $(1\times3)$ and the vertical direction is processed at $(3\times1)$.

\subsubsection{Network structure}\label{Network structure}
\begin{figure}[htbp]
	\centerline{\includegraphics[width=13cm]{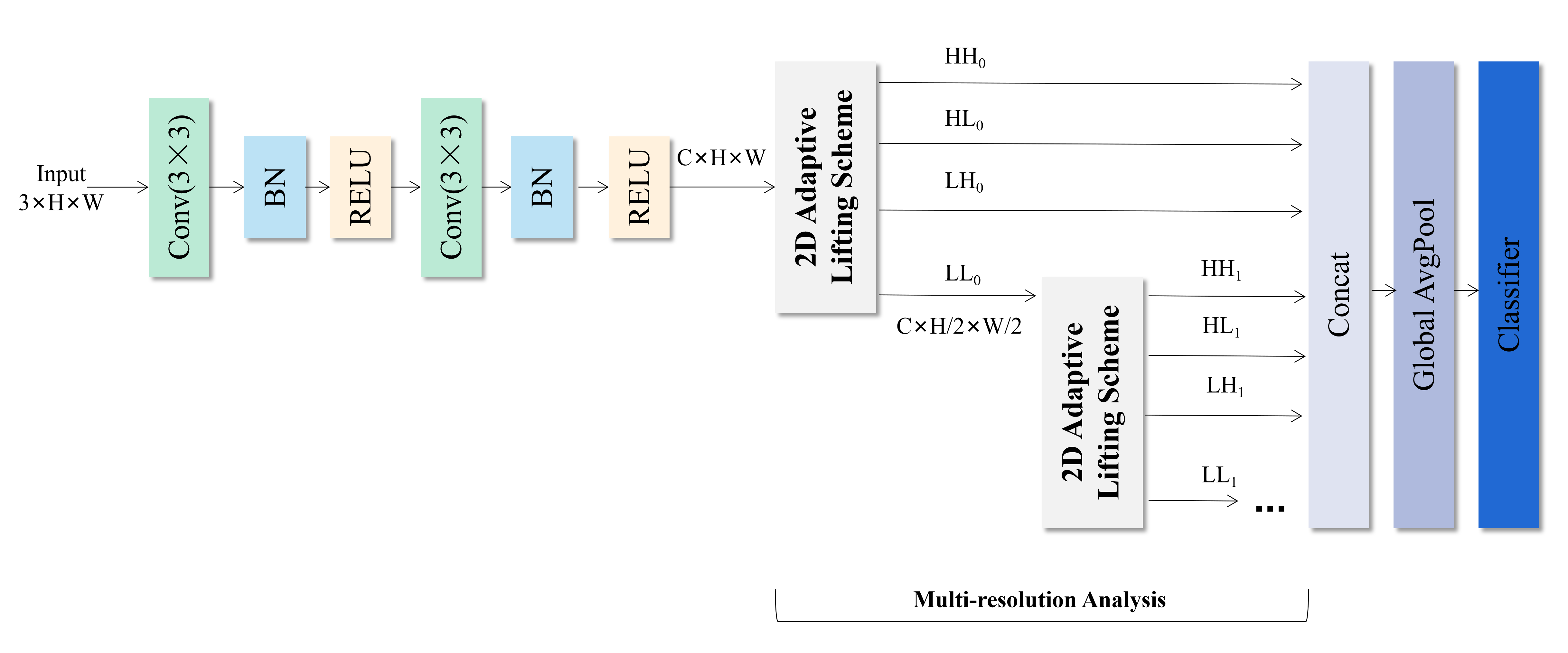}}
	\vspace*{8pt}
	\caption{Network structure of DAWN.}
	\label{fig:3}
\end{figure}

Fig. \ref{fig:3} illustrates the network structure of DAWN. The number of wavelet decomposition levels, which corresponds to the number of 2D Adaptive Lifting Scheme shown in the figure, is an optional hyperparameter with an upper bound. The maximum value $N$ of wavelet decomposition levels is $\lfloor \log_2W - \log_24 \rfloor$ when the size of the input features is $W \times W$.

Maria et al.\cite{1} achieve multi-resolution analysis by concatenating various modules through the $LL$ component of the output of the previous 2D Adaptive Lifting Scheme module being used as the input of the next module. Furthermore, they included a CNN of depth 2 before the multi-resolution analysis group to generate a discriminative representation of the data required for texture classification. At the end of the network, the detailed sub-bands of all Lifting Scheme modules are concatenated with the approximate sub-bands of the last Lifting Scheme module and inputted into the classifier for texture classification.

It is noteworthy that the wavelet module and the CNN learn features in the frequency and spatial domains, respectively. Nevertheless, the design approach of DAWN is to concatenate these two and incorporate only a shallow CNN which does not leverage the full potential of the CNN in learning features in the spatial domain.

\subsection{Design in our network}\label{Design in our network}

\subsubsection{Adaptive wavelets transform module}\label{Adaptive wavelets transform module}
\begin{figure}[htbp]
	\centerline{\includegraphics[width=13cm]{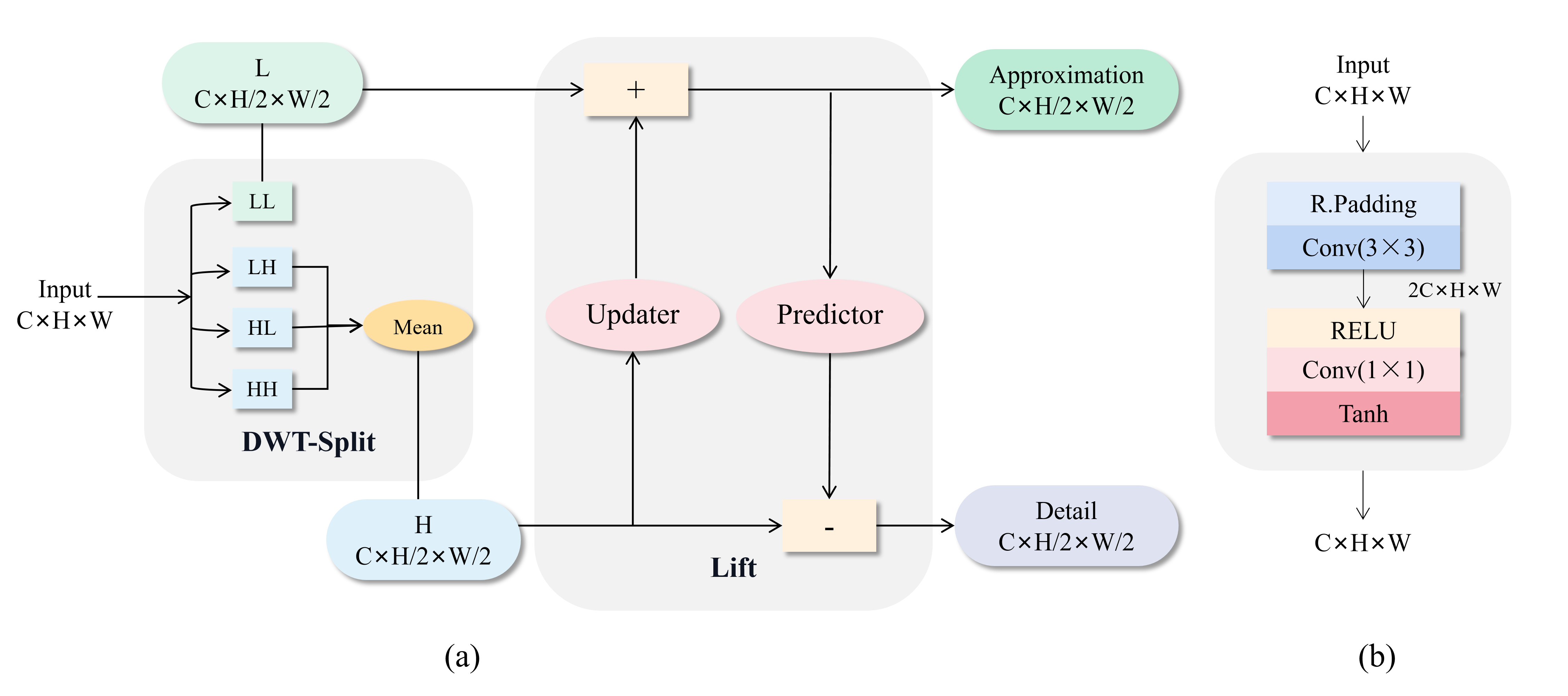}}
	\vspace*{8pt}
	\caption{(a) Structure of the AWTM. (b) Structure of the updater and predictor within AWTM.}
	\label{fig:4}
\end{figure}

The Adaptive Wavelets Transform Module (AWTM) is further designed on the basis of the wavelet module of DAWN, and its structure is shown in Fig. \ref{fig:4}.

\textbf{Split} Sampling lattice splitting and discrete wavelet transform splitting (DWT-Split) are both effective means of Split. \cite{13} The designed network utilizes the DWT-Split method and applies the Haar wavelet transformation. In this method, the input signal of size $C\times H\times W$ decomposes into four components of size $C\times H/2\times W/2$ at the split stage. Component $LL$ is sent to the lift stage, while the other three high-frequency components are averaged as input $H$ in the lift stage. By doing so, the method can achieve wavelet transform processing of two-dimensional signals through a single Lifting Scheme module, which reduces the number of parameters required for training.

\textbf{Updater and Predictor} Since the network in this paper processes the signal in two dimensions simultaneously, the convolutional kernel size of the first convolutional layer in the updater and predictor is set to $(3\times3)$.

\textbf{Loss} Same as Ref.~\refcite{1}, we set the loss function shown in ``(\ref{eq:2.2})'' for the wavelet module. $\alpha$ and $\beta$ are hyperparameters that regulate the weights of the two regular terms. $H$ denotes the Huber's parametrization, $D_{i} $ denotes the detail coefficients of the output of the wavelet decomposition at level $i$. $m^{I}_{i}$ and $m^{A}_{i} $ denote the average of the input, approximation coefficients of the output of the wavelet decomposition at level $i$, respectively. The first regular term minimizes the sum of Huber parametrization of the detail coefficients on all decomposition levels, and the second regular term minimizes the sum of $L2$ norm of the difference between $m^{I}_{i} $ and $m^{A}_{i} $ on all decomposition levels.
\begin{equation}\label{eq:2.2}
	Loss_{WT}=\alpha{\sum^L_{i=1} {H(D_{i})}}+\beta{\sum^L_{i=1} {\Vert{m^{I}_{i}-m^{A}_{i}}\Vert^{2}_{2}} }.
\end{equation}

\subsubsection{Network struture}\label{Our network struture}
\begin{figure}[htbp]
	\centerline{\includegraphics[width=13cm]{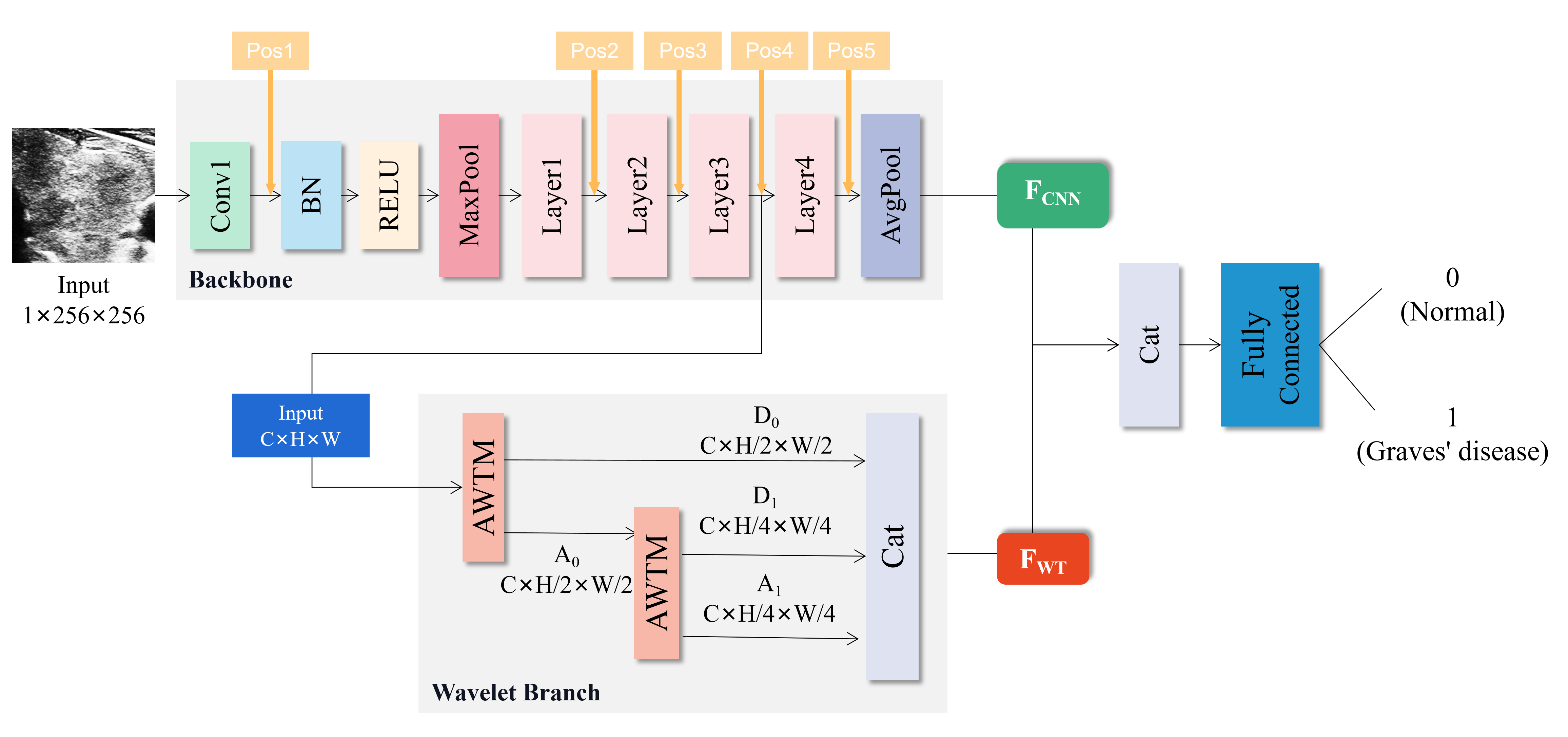}}
	\vspace*{8pt}
	\caption{Network structure of ResNet$\_$WT.}
	\label{fig:5}
\end{figure}

The designed network structure is shown in Fig. \ref{fig:5}. According to the experimental results in Sec. \ref{Ablation experiments and hyperparameter studies}, the optimal insertion position (Pos4) and decomposition level (2) were selected for the drawing of Fig. \ref{fig:5}. Studies have shown that shallower CNNs are more suitable for ultrasonic texture analysis tasks than deeper CNNs.\cite{15} Accordingly, we chose ResNet18 as our backbone.

The ultrasonic grayscale texture image is input into the network, which consists of two parts: the backbone network ResNet18 and the wavelet branch. Different from DAWN, this paper accesses the wavelet branch in parallel after an insertion position in the backbone network to learn the frequency domain characteristics separately. At the end of the network, the spatial domain feature $F_{CNN}$ from backbone network and the frequency domain feature $F_{WT}$ from wavelet branch are concatenated together and input to the fully connected layer for diagnosis.

\textbf{Number of decomposition levels} This is an optional hyperparameter in the method, which is expressed within the network as the number of AWTM within the wavelet branch. If the user does not set this parameter, the network will default to using the highest decomposition levels $N$ that the input image can perform (as described in Sec. \ref{Network structure}), for the reasons described in Sec. \ref{Ablation experiments and hyperparameter studies}. 

\textbf{Multi-resolution Analysis} The multi resolution analysis integrated in this article's network is shown in the Wavelet Branch in Fig. \ref{fig:5}, which shows the situation when the wavelet decomposition level is set to 2. Unlike DAWN's design, the multi-resolution analysis module is inserted in parallel into the backbone network at a certain insertion position, rather than concatenated to a shallow CNN.

\textbf{Insertion locations} This is one of the few hyperparameters in the designed network. The proposed method selects five insertion positions within ResNet18 that are not simultaneously present. The performance of the different insertion locations will be described in Sec. \ref{Ablation experiments and hyperparameter studies}.

\textbf{Loss} Same as Ref.~\refcite{1}, we set the loss function shown in ``(\ref{eq:2.3})'' for the designed network, where $g_{i}$ and $p_{i}$ denotes the label value of class $i$ and the probability that the input image predicted by the network belongs to that class. The first term in the loss function is the cross-entropy loss of the classification, $Loss_{WT}$ is the wavelet module loss function described in Sec. \ref{Adaptive wavelets transform module}.
\begin{equation}\label{eq:2.3}
	Loss=-\sum^C_{i=1} {g_{i}\log (p_{i})}+ Loss_{WT}.
\end{equation}

\section{Experiments}\label{Experiments}
\subsection{Data set and pre-processing}\label{Data set and pre-processing}
\textbf{Thyroid ultrasound dataset} Region of Interest (ROI) extraction was performed according to the physician's instruction, and data augmentation of horizontal flip, affine transformation, and rotation was done on the data to balance the amount of training data of different categories to prevent the network from overfitting, and histogram equalization was done on all data to reduce the influence of image contrast on the diagnosis. The data set is shown in Table \ref{table:1}.

\begin{table}[ht]
	\tbl{The data set after data augmentation.\label{table:1}}
	{\begin{tabular}{@{}cccc@{}} \toprule
			Data set & Normal & Graves' disease &
			Total amount \\ \colrule
			Train\hphantom{00} & \hphantom{0}243 & \hphantom{0}259 & 502 \\
			Test\hphantom{00} & \hphantom{0}42 & \hphantom{0}58 & 100 \\ \botrule
	\end{tabular}}
\end{table}

\begin{figure}[htbp]
	\centerline{\includegraphics[width=13cm]{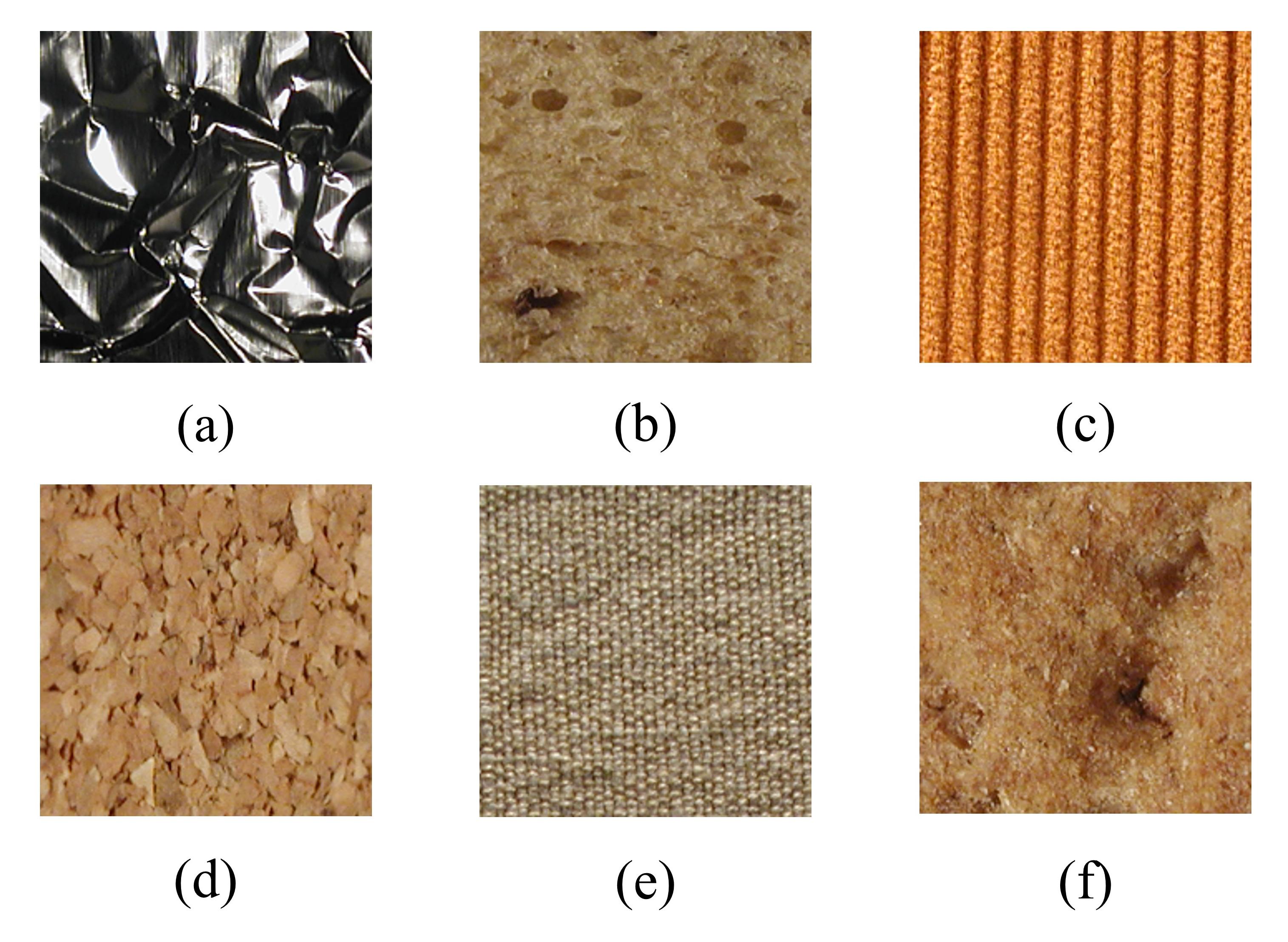}}
	\vspace*{8pt}
	\caption{Sample images of some classes in the KTH dataset. (a)aluminium foil; (b) brown bread; (c) corduroy; (d) cork; (e) cotton; (f) cracker.}
	\label{fig:6}
\end{figure}

\textbf{KTH-TIPS2-B} This public database was developed by the Computational Vision and Active Perception Laboratory (CVAP) of KTH Royal Institute of Technology. There are three versions available: KTH-TIPS, KTH-TIPS2-A, and KTH-TIPS2-B. In this study, we selected the most widely used KTH-TIPS2-B. This version contains 11 classes, each with 4 samples and 108 images per sample. We selected one sample from each class for training and tested the remaining samples, resulting in four data segmentation methods. Each data split method has 950 training images, 238 validation images, and 3564 test images. We resized all images to 256 $\times$ 256 and divided the training and validation sets in a 4:1 ratio within the training set. Data augmentation included applying random mirroring, rotation, brightness adjustment, and other operations. Fig. \ref{fig:6} shows example images of some classes in the KTH dataset.

\subsection{Parameter settings}\label{Parameter settings}
The ROI data processed using histogram equilibrium are fed into the network, and the input channels of the network are all set to 1. The SGD optimizer is used in training, the momentum is set to 0.9, and each network is trained with 100 epochs. According to the study in Ref.~\refcite{1}, $\alpha$ and $\beta$ in the loss function are taken as 0.1. The batch size is set to 8, and the learning rate adjustment mechanism of decaying half the learning rate every 10 epochs is used. The initial learning rate is taken as $3 \times 10^{2}$ for DAWN and $1 \times 10^{-3}$ for the rest of the networks. In the training of ultrasound images, for DAWN and VGG, the input image size is set to $224 \times 224$, and the input image size is set to 256 $\times$ 256 for the rest of the networks, all input channels of the network are 1. In the training of the KTH dataset, the input image size of all networks is 256 $\times$ 256, and the input channel is 3. The DenseNet growth factor is taken as 32, and the network depth is taken as 121. All networks are trained from scratch.

\subsection{Evaluation indexes}\label{Evaluation indexes}
The accuracy and recall are adopted as the evaluation indexes of the method. The confusion matrix is shown in Table \ref{table:2}.

\begin{table}[ht]
	\tbl{Confusion matrix.\label{table:2}}
	{\begin{tabular}{@{}cccc@{}} \toprule
			\diagbox{Predict}{Ground truth} & Positive & Negative \\ \colrule
			Positive\hphantom{00} & \hphantom{0}TP & \hphantom{0}FP\\
			Negative\hphantom{00} & \hphantom{0}FN & \hphantom{0}TN\\ \botrule
	\end{tabular}}
\end{table}

Accuracy represents the percentage of correct predictions among all samples and is calculated as ``(\ref{eq:3.1})''.
\begin{equation}\label{eq:3.1}
	Accuracy=\dfrac{TP+TN}{TP+FP+FN+TN}.
\end{equation}

Recall represents the percentage of correct predictions among all positive samples (Graves' disease) and is calculated as ``(\ref{eq:3.2})''.
\begin{equation}\label{eq:3.2}
	Recall=\dfrac{TP}{TP+FN}.
\end{equation}

\subsection{Comparison with other networks}\label{Comparison with other networks}
\begin{figure}[htbp]
	\centerline{\includegraphics[width=8cm]{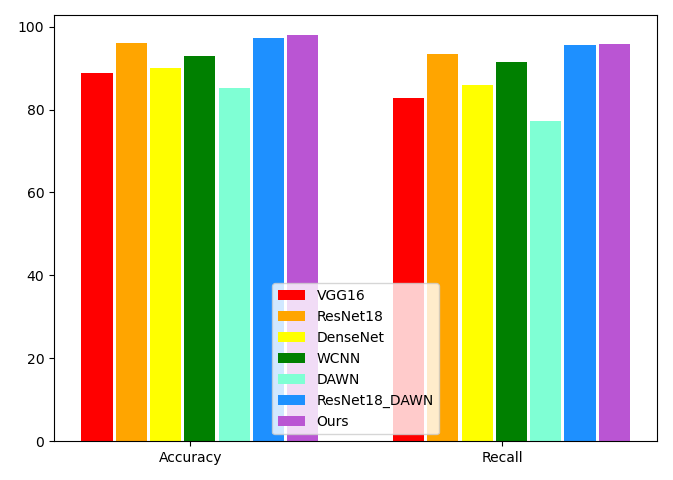}}
	\vspace*{8pt}
	\caption{Bar chart of accuracy and recall in the experimental results.}
	\label{fig:7}
\end{figure}

\begin{figure}[htbp]
	\centerline{\includegraphics[width=9cm]{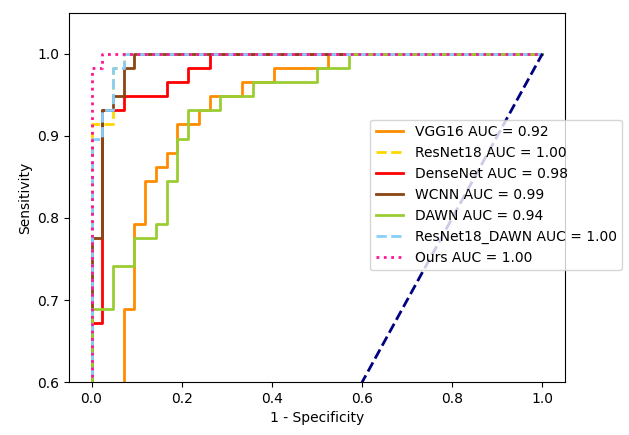}}
	\vspace*{8pt}
	\caption{ROC curves of ultrasound texture classification results of Graves' disease for each network.}
	\label{fig:8}
\end{figure}

\textbf{Thyroid ultrasound dataset} We selected VGG,\cite{11} ResNet,\cite{5} DenseNet,\cite{6} WCNN,\cite{4} and DAWN\cite{1} as control methods. The first three networks are commonly used classification networks, WCNN and DAWN, which fuse wavelet transform into CNN, are the focus of comparison with the proposed method. Since the input image size of DAWN is 224, the decomposition level of its wavelet transform is taken as 5. According to the parameter setting of Sec. \ref{Parameter settings}, the data with the number of channels as 1 are input into the network, respectively, and the texture binary classification is performed by training 100 epochs from scratch. Taking the average accuracy of 10 inferences as the consideration indicator. The results are shown in Table \ref{table:3} and Fig. \ref{fig:7}, and the ROC graph is shown in Fig. \ref{fig:8}.

\begin{table}[ht]
	\tbl{Ultrasonic texture classification results of Graves' disease for each network.\label{table:3}}
	{\begin{tabular}{@{}cccc@{}} \toprule
			Network & Accuracy($\%$) & Recall($\%$) \\ \colrule
			VGG16\hphantom{00} & \hphantom{0}88.838 & \hphantom{0}82.845\\
			ResNet18\hphantom{00} & \hphantom{0}\textbf{95.969} & \hphantom{0}\textbf{93.448}\\ 
			DenseNet\hphantom{00} & \hphantom{0}90.043 & \hphantom{0}86.034\\ 
			WCNN\hphantom{00} & \hphantom{0}93.000 & \hphantom{0}91.379\\ 
			DAWN\hphantom{00} & \hphantom{0}85.100 & \hphantom{0}77.241\\ \colrule
			ResNet18$\_$DAWN\hphantom{00} & \hphantom{0}97.270 & \hphantom{0}95.603\\
			ResNet18$\_$WT(Ours)\hphantom{00} & \hphantom{0}\textbf{97.900} & \hphantom{0}\textbf{95.862}\\
			\botrule
	\end{tabular}}
\end{table}


Based on the experimental results, VGG was unable to achieve satisfactory accuracy with the same training settings due to the large number of required training parameters. Although DenseNet and ResNet18 both have skip connection structures, ResNet18's shallower network structure makes it more suitable for the ultrasonic texture classification task, yielding superior performance. Conversely, DAWN's performance in this task was poor, demonstrating that a single frequency domain feature cannot achieve high accuracy. On the other hand, WCNN outperformed DAWN, ranking second best among the selected CNNs, validating the effectiveness of using both frequency and spatial domain information. 

ResNet18$\_$WT, network designed in this paper, showcases the optimal performance. Compared with optimal CNN (ResNet18), it has improved accuracy and recall by 1.931$\%$ and 2.414$\%$, respectively. In the ROC curve, this study zooms in on the upper-left corner of the original curve, which clearly shows that the network developed in this study has the largest Area Under the Curve (AUC), demonstrating the optimal classification performance. Furthermore, the maximum improvement values of the proposed method over the rest of the methods in terms of accuracy and recall are shown in Table \ref{table:4}. It can be seen that our network achieves remarkable accuracy improvement over the classical classification network and existing methods in the same field.

\begin{table}[ht]
	\tbl{Maximum improvement in accuracy of ResNet18$\_$WT over each CNN.\label{table:4}}
	{\begin{tabular}{@{}ccccccc@{}} \toprule
			\makecell[c]{Maximum \\ improvement \\ ($\%$)}  & VGG16 & ResNet18 & DenseNet &  WCNN & DAWN & ResNet18$\_$DAWN\\ \colrule
			Accuracy\hphantom{00} & \hphantom{0}9.062 & \hphantom{0}1.931 & \hphantom{0}7.857 & \hphantom{0}4.900 & \hphantom{0}12.800 & \hphantom{0}0.630 \\
			Recall\hphantom{00} & \hphantom{0}13.017 & \hphantom{0}2.414 & \hphantom{0}9.828 & \hphantom{0}4.483 & \hphantom{0}18.621 & \hphantom{0}0.259 \\ \botrule
	\end{tabular}
}
\end{table}

\begin{figure}[htbp]
	\centerline{\includegraphics[width=12cm]{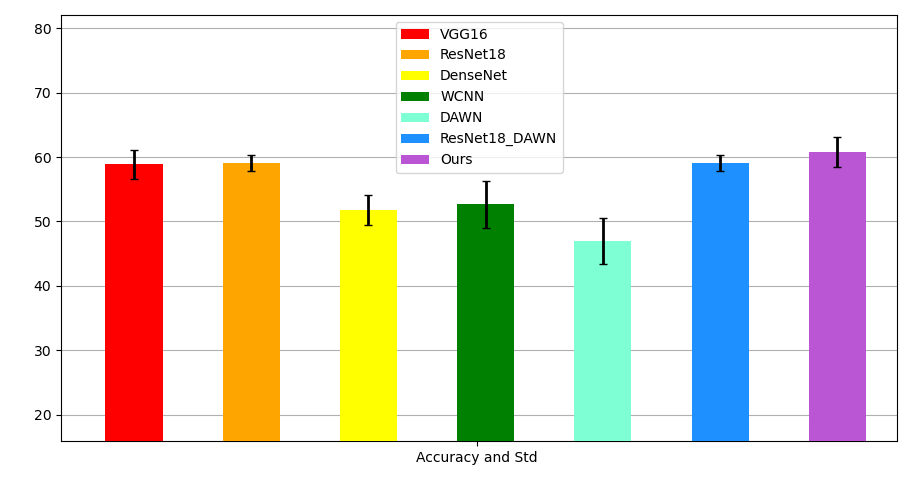}}
	\vspace*{8pt}
	\caption{The accuracy and standard deviation of classification experiments conducted by various networks on the KTH dataset.}
	\label{fig:9}
\end{figure}

\textbf{KTH-TIPS-B} We selected same networks for comparative experiments, and used the accuracy of network testing under four split datasets as the evaluation indicator. We compared the classification results of several networks to verify the performance of the proposed method on natural texture images. The experimental results are shown in Table \ref{table:5} and Fig. \ref{fig:9}. It can be seen that ResNet18 is still the CNN with the highest accuracy among the comparison methods. The average accuracy of ResNet18-DAWN did not exceed ResNet18, but the average accuracy of ResNet18$\_$WT exceeded the optimal baseline network, which fully demonstrates the effectiveness of the AWTM design in this article. AWTM not only fully learns the frequency domain features of ultrasound data, but also has stronger adaptability to natural image datasets than the wavelet module in DAWN, showcased the significant promotion potential of the AWTM module. At the same time, our network has improved the average accuracy by 1.669\% on the basis of ResNet18, which once again proves the effectiveness of the parallel wavelet branch network design in this article.

\begin{table}[ht]
	\tbl{Classification results of KTH datasets for each network.\label{table:5}}
	{\begin{tabular}{@{}cc@{}} \toprule
			Network & Accuracy $\pm$ Std \\ \colrule
			VGG16\hphantom{00} & \hphantom{0}56.571$\pm$2.3 \\
			ResNet18\hphantom{00} & \hphantom{0}\textbf{59.096$\pm$1.2}\\ 
			DenseNet\hphantom{00} & \hphantom{0}51.809$\pm$2.3 \\ 
			WCNN\hphantom{00} & \hphantom{0}52.639$\pm$3.7 \\ 
			DAWN\hphantom{00} & \hphantom{0}47.003$\pm$3.6 \\ \colrule
			ResNet18$\_$DAWN\hphantom{00} & \hphantom{0}59.057$\pm$1.3 \\
			ResNet18$\_$WT(Ours)\hphantom{00} & \hphantom{0}\textbf{60.765$\pm$2.3} \\
			\botrule
	\end{tabular}}
\end{table}

\subsection{Ablation experiments and hyperparameter studies}\label{Ablation experiments and hyperparameter studies}
The last two CNN variants shown in Table \ref{table:3} incorporate ideas from this paper. ResNet18$\_$DAWN adapts the proposed method by paralleling the wavelet module in DAWN to the five positions indicated in Fig. \ref{fig:5}. It can be observed that this method achieved significant accuracy improvement over the baseline ResNet18, achieving an augmented accuracy and recall by 1.301$\%$ and 2.155$\%$, respectively. This result emphasizes the effectiveness of paralleling adaptive wavelet modules to CNNs to learn spatial and frequency domain features.

ResNet18$\_$WT, the network designed in this paper, enhances ResNet18$\_$DAW-N's wavelet module by substituting the Split method with Haar discrete wavelet transform. This modification decreases the number of lifting scheme modules required for one level of wavelet decomposition from 3 to 1, which results in a significant reduction (more than 30,000 each level) in the volume of trainable parameters. The experimental results presented in Table \ref{table:3} demonstrate that the proposed module design yields more accurate predictions than ResNet18$\_$DAWN. This indicates that our AWTM design method is effective.

\begin{figure}[htbp]
	\centerline{\includegraphics[width=12cm]{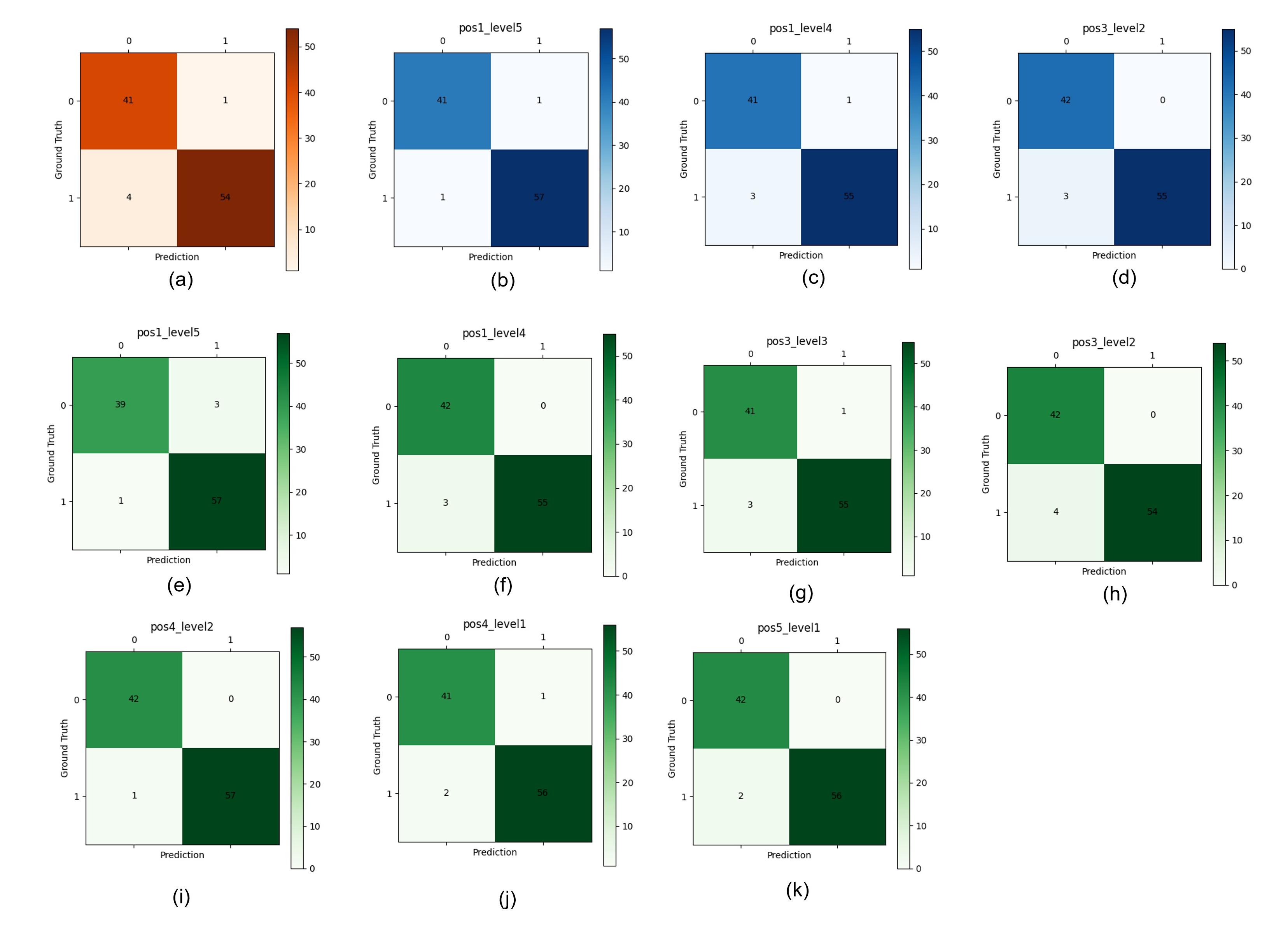}}
	\vspace*{8pt}
	\caption{(a) Confusion matrix for the inference stage of ResNet18; (b)-(d) Confusion matrix of ResNet18\_DAWN in the inference phase with different hyperparameter settings; (e)-(k) Confusion matrix of ResNet18\_WT in the inference phase with different hyperparameter settings.}
	\label{fig:10}
\end{figure}

To further demonstrate the effectiveness of the proposed method, we compared ResNet18$\_$DAWN and ResNet18$\_$WT concerning the hyperparameters (insertion position and decomposition level) on ultrasound dataset, as shown in Table \ref{table:6}. Fig. \ref{fig:10} shows the confusion matrix of ResNet and the two designed networks in the inference phase. 
\begin{table}[ht]
	\tbl{Experimental results of hyperparameters of the designed network on ultrasound dataset.\label{table:6}}
	{\begin{tabular}{@{}ccccc@{}} \toprule
			Network  & Insertion position & Decomposition level & Accuracy($\%$) &  Recall($\%$) \\ \colrule
			\multirow{3}{*}{ResNet18$\_$DAWN}\hphantom{00} & \hphantom{0}pos1 & \hphantom{0}5 & \hphantom{0}\it97.083 & \hphantom{0}\it95.086\\
														&	\hphantom{0}pos1 & \hphantom{0}4 & \hphantom{0}\textbf{\it97.270} & \hphantom{0}\textbf{\it95.603}\\
														&	\hphantom{0}pos3 & \hphantom{0}2 & \hphantom{0}\it96.050 & \hphantom{0}\it93.621\\ 
			\multicolumn{3}{c}{Average value of performance improvement compared to ResNet18} & \hphantom{0}0.832 & \hphantom{0}1.322 \\ \colrule
			\multirow{7}{*}{ResNet18$\_$WT(Ours)}\hphantom{00} & \hphantom{0}pos1 & \hphantom{0}5 & \hphantom{0}\it97.513 & \hphantom{0}\textbf{\it96.121}\\
															&	\hphantom{0}pos1 & \hphantom{0}4 & \hphantom{0}\it96.804 & \hphantom{0}\it95.172\\ 
															&	\hphantom{0}pos3 & \hphantom{0}3 & \hphantom{0}\it96.105 & \hphantom{0}\it94.138\\ 
															&	\hphantom{0}pos3 & \hphantom{0}2 & \hphantom{0}95.111 & \hphantom{0}\it95.603\\ 
															&	\hphantom{0}pos4 & \hphantom{0}2 & \hphantom{0}\textbf{\it97.900} & \hphantom{0}\it95.862\\ 
															&	\hphantom{0}pos4 & \hphantom{0}1 & \hphantom{0}\it97.277 & \hphantom{0}\it95.431\\ 
															&	\hphantom{0}pos5 & \hphantom{0}1 & \hphantom{0}\it97.660 & \hphantom{0}\it95.603\\ 
			\multicolumn{3}{c}{Average value of performance improvement compared to ResNet18} & \hphantom{0}0.941$\uparrow$ & \hphantom{0}1.971$\uparrow$ \\
														\botrule
	\end{tabular}}
\begin{tabnote}
	The results displayed in italics indicate metrics performing better than ResNet18, and the bolded data represent the highest values of corresponding metrics within the network.
\end{tabnote}
\end{table}

Experiments have shown that in ResNet18$\_$DAWN, the accuracy of only 2 insertion positions and 3 decomposition levels outperforms ResNet18. Under the proposed method ResNet18$\_$WT, the number of hyperparameter settings with excellent accuracy performance increased, from 3 in ResNet18$\_$DAWN to 7. At the same time, compared to ResNet18$\_$DAWN, the highest values of accuracy and recall of the proposed method are further improved by 0.630$\%$ and 0.518$\%$, respectively. 

In addition, the average accuracy of all effective settings in our network has also been improved, compared with ResNet18, the accuracy and recall improved by 0.941$\%$ and 1.971$\%$ on average, respectively. The improvement of both indicators is greater than that of ResNet18$\_$DAWN, confirming the effectiveness of our AWTM design method.

Similarly, we conducted ablation experiments and hyperparameter analysis on the natural dataset KTH, and the results are shown in Table \ref{table:7}. We demonstrated the classification accuracy of each hyperparameter setting for ResNet18, ResNet18$\_$DAWN, and ResNet18$\_$WT under four data partitioning scenarios on the KTH dataset.
ResNet18$\_$DAWN has a total of 7 hyperparameter settings that outperform ResNet18, while  ResNet18$\_$WT has a total of 24 hyperparameter settings that exceed the baseline ResNet18 in all split modes, which strongly demonstrates the effectiveness of our network framework design. At the same time, ResNet18$\_$WT's effective number of hyperparameters and highest accuracy in each split mode far exceed ResNet18$\_$DAWN, which once again proves the effectiveness of our design for AWTM. Experimental results have shown that our network exhibits significant accuracy advantages on both ultrasound datasets and natural image sets.

\begin{table}[ht]
	\tbl{Experimental results of hyperparameters of the designed network on KTH-TIPS-B.\label{table:7}}
	{\begin{tabular}{@{}ccccccc@{}} \toprule
			Network  &  \makecell[c]{Insertion\\ position} & \makecell[c]{Decompo-\\sition\\ level} & \makecell[c]{Train\\ Sample$\_$A} & \makecell[c]{Train\\ Sample$\_$B} &  \makecell[c]{Train\\ Sample$\_$C} &  \makecell[c]{Train\\ Sample$\_$D} \\ \colrule
			
			ResNet18\hphantom{00} &
			\hphantom{0}$-$ & \hphantom{0}$-$ & \hphantom{00} 60.101&
			\hphantom{00} 60.449&
			\hphantom{00} 57.969&
			\hphantom{00} 57.865 \\ \colrule
			
			\multirow{9}{*}{\makecell[c]{ResNet18\\$\_$DAWN}}\hphantom{00} & 
			\hphantom{0}1 & 
			\hphantom{0}5 & 
			\hphantom{00} \it61.004&
			\hphantom{00} 57.497&
			\hphantom{00} 57.270&
			\hphantom{00} 56.322 \\&
			\hphantom{0}1 & 
			\hphantom{0}4 & 
			\hphantom{00} \it60.777&
			\hphantom{00} \textbf{59.731}&
			\hphantom{00} \textbf{57.287}&
			\hphantom{00} \textbf{\it58.432} \\&
			\hphantom{0}2 & 
			\hphantom{0}4 & 
			\hphantom{00} 58.384&
			\hphantom{00} 54.675&
			\hphantom{00} 54.248&
			\hphantom{00} 57.225 \\&
			\hphantom{0}2 & 
			\hphantom{0}3 & 
			\hphantom{00} 59.116&
			\hphantom{00} 55.707&
			\hphantom{00} 52.980&
			\hphantom{00} 57.803 \\&
			\hphantom{0}3 & 
			\hphantom{0}3 & 
			\hphantom{00} \it60.727&
			\hphantom{00} 54.570&
			\hphantom{00} 52.988&
			\hphantom{00} 57.312 \\&
			\hphantom{0}3 & 
			\hphantom{0}2 & 
			\hphantom{00} \it60.379&
			\hphantom{00} 56.131&
			\hphantom{00} 54.231&
			\hphantom{00} 56.504 \\&
			\hphantom{0}4 & 
			\hphantom{0}2 & 
			\hphantom{00} 59.212&
			\hphantom{00} 58.824&
			\hphantom{00} 54.697&
			\hphantom{00} 55.856 \\&
			\hphantom{0}4 & 
			\hphantom{0}1 & 
			\hphantom{00} \textbf{\it61.496}&
			\hphantom{00} 58.855&
			\hphantom{00} 57.020&
			\hphantom{00} 57.489 \\&
			\hphantom{0}5 & 
			\hphantom{0}1 & 
			\hphantom{00} \it61.049&
			\hphantom{00} 55.867&
			\hphantom{00} 54.927&
			\hphantom{00} 54.753 \\ \colrule
			
			\multirow{9}{*}{\makecell[c]{ResNet18\\$\_$WT(Ours)}}\hphantom{00} & 
			\hphantom{0}1 & 
			\hphantom{0}5 & 
			\hphantom{00} 59.800&
			\hphantom{00} 57.217&
			\hphantom{00} 57.304&
			\hphantom{00} \it59.172 \\&
			
			\hphantom{0}1 & 
			\hphantom{0}4 & 
			\hphantom{00} 58.800&
			\hphantom{00} 57.817&
			\hphantom{00} \it58.361&
			\hphantom{00} \textbf{\it59.332} \\&
			
			\hphantom{0}2 & 
			\hphantom{0}4 & 
			\hphantom{00} \it62.000&
			\hphantom{00} \it61.299&
			\hphantom{00} \it58.423&
			\hphantom{00} \it58.709 \\&
			
			\hphantom{0}2 & 
			\hphantom{0}3 & 
			\hphantom{00} 
			\it62.177&
			\hphantom{00} \it58.939&
			\hphantom{00} \it59.150&
			\hphantom{00} \it59.285 \\&
			
			\hphantom{0}3 & 
			\hphantom{0}3 & 
			\hphantom{00} \it60.738&
			\hphantom{00} \it61.156&
			\hphantom{00} 57.093&
			\hphantom{00} \it58.235 \\&
			
			\hphantom{0}3 & 
			\hphantom{0}2 & 
			\hphantom{00} \it63.117&
			\hphantom{00} \textbf{\it62.652}&
			\hphantom{00} 56.829&
			\hphantom{00} \it57.983 \\&
			
			\hphantom{0}4 & 
			\hphantom{0}2 & 
			\hphantom{00} \it61.378&
			\hphantom{00} 59.874&
			\hphantom{00} \it58.249&
			\hphantom{00} 57.068 \\&
			
			\hphantom{0}4 & 
			\hphantom{0}1 & 
			\hphantom{00} \it62.085&
			\hphantom{00} 59.515&
			\hphantom{00} \textbf{\it60.331}&
			\hphantom{00} \it58.137 \\&
			
			\hphantom{0}5 & 
			\hphantom{0}1 & 
			\hphantom{00} \textbf{\it64.565}&
			\hphantom{00} 59.997&
			\hphantom{00} \it59.913&
			\hphantom{00} \it58.583 \\ \botrule
	\end{tabular}}
	\begin{tabnote}
		The italicized data indicates that under this setting, the designed network has a higher accuracy than ResNet18 trained on the same sample, The bold data represents the highest accuracy of the current network within this split method.
	\end{tabnote}
\end{table}

Our experiments also investigate the impact of the two hyperparameters, insertion position and decomposition level, on the performance of the proposed method. The data from ResNet18$\_$WT in Table \ref{table:6} demonstrate that the impact of insertion position on accuracy is insubstantial in the ultrasound dataset, which suggests that the proposed method can use both simple features at shallow level, and domain-specific features at deep level for texture classification. We said in the previous section that the decomposition level is obtained as large as possible by default when the user does not set it, and the experimental data confirm the correctness of our design. As shown in the data of ResNet18$\_$WT in Table \ref{table:6}, where the accuracy is highest at each insertion position when the decomposition level is taken as maximum. The texture of natural image datasets is more complex, so it can be seen from Table \ref{table:7} that the two hyperparameters have a greater impact on model accuracy. However, it is stable that the highest accuracy of our model is always higher than the baseline network ResNet18.

\section{Conclusion}\label{Conclusion}
We utilized a convolutional neural network that incorporates wavelet transform to achieve accurate diagnosis of ultrasound images of Graves' disease. Our network uses Lifting Scheme to implement an adaptive wavelet transform module and parallels a learnable wavelet branch integrated with multi-resolution analysis to CNN to learn features in both spatial and frequency domains. Our approach outperforms existing correlation methods and advanced CNNs in accuracy while being more accessible to physicians due to its interpretability. Although our approach is focused on Graves' disease research, the classification results on the KTH dataset show that our approach has the potential to be extended to other tasks of texture analysis. In the future, we will further investigate the generalizability and robustness of the model to improve its performance on diverse types of datasets.

\section*{Acknowledgments}
This work was supported by Chongaina Municipal Health Commission
Award (No. 2020MSXM088), Chongqing Municipal Health Commission
Award (No. 2018GDRCO06) and the Fundamental Research Funds for Army Medical University (No. 2021HQZX08, No. LJ20222Z060052).

\end{document}